\documentclass[10pt]{article}

\usepackage{amsmath,amsfonts,bm}
\usepackage{hyperref}
\usepackage{url}
\usepackage{todonotes}
\usepackage{nicefrac}
\usepackage{booktabs}
\usepackage{cleveref}
\usepackage[whole]{bxcjkjatype}
\usepackage{graphicx}
\usepackage{enumitem}
\usepackage{pifont}
\usepackage{natbib}
\usepackage{authblk}
\usepackage{rotating}
\usepackage{tabularx}
\usepackage{ltablex}
\usepackage{orcidlink}

\usepackage[margin=1in]{geometry}

\title{Stabilizing Reasoning in Medical LLMs with Continued Pretraining and Reasoning Preference Optimization}

\author[1,2]{Wataru~Kawakami\,\orcidlink{0009-0006-3785-3543}\footnote{This work was done when W.K. worked at Preferred Networks as a part-time engineer.}}
\author[1]{Keita Suzuki}
\author[1]{Junichiro Iwasawa\,\orcidlink{0000-0002-2560-5650}\footnote{iwasawa@preferred.jp}}
\affil[1]{Preferred Networks Inc., Tokyo, Japan}
\affil[2]{Graduate School of Information Science and Technology, The University of Tokyo, Tokyo, Japan}
\date{}

\begin{document}
\maketitle
\begin{abstract}
Large Language Models (LLMs) show potential in medicine, yet clinical adoption is hindered by concerns over factual accuracy, language-specific limitations (e.g., Japanese), and critically, their reliability when required to generate reasoning explanations---a prerequisite for trust. 
This paper introduces Preferred-MedLLM-Qwen-72B, a 72B-parameter model optimized for the Japanese medical domain to achieve both high accuracy and stable reasoning.
We employ a two-stage fine-tuning process on the Qwen2.5-72B base model: first, Continued Pretraining (CPT) on a comprehensive Japanese medical corpus instills deep domain knowledge. 
Second, Reasoning Preference Optimization (RPO), a preference-based method, enhances the generation of reliable reasoning pathways while preserving high answer accuracy.
Evaluations on the Japanese Medical Licensing Exam benchmark (IgakuQA) show Preferred-MedLLM-Qwen-72B achieves state-of-the-art performance (0.868 accuracy), surpassing strong proprietary models like GPT-4o (0.866). 
Crucially, unlike baseline or CPT-only models which exhibit significant accuracy degradation (up to 11.5\% and 3.8\% respectively on IgakuQA) when prompted for explanations, our model maintains its high accuracy (0.868) under such conditions.
This highlights RPO's effectiveness in stabilizing reasoning generation. 
This work underscores the importance of optimizing for reliable explanations alongside accuracy. 
We release the Preferred-MedLLM-Qwen-72B model weights to foster research into trustworthy LLMs for specialized, high-stakes applications.
\end{abstract}

\section{Introduction}
The rapid advancement of Large Language Models (LLMs) marks a significant milestone in artificial intelligence, showcasing impressive abilities in understanding, synthesizing, and reasoning with complex information across various domains~\citep{hurst2024gpt, grattafiori2024llama, yang2024qwen2, abe2024plamo}. 
In medicine, these models hold immense potential, capable of achieving expert-level performance on benchmarks~\citep{nori2024medprompto1,singhal2025medpalm} and offering utility in applications such as clinical text summarization~\citep{van2024adapted}, patient interaction~\citep{cosentino2024towards}, diagnostic decision support~\citep{tu2025towards} and information extraction~\citep{mou2024radlink}. 
Foundation models, especially those developed for English, have pushed performance boundaries in medical domains~\citep{saab2024medgemini,brodeur2024}.

However, the deployment of LLMs in high-stakes clinical settings faces substantial obstacles. 
Factual inaccuracies (hallucinations) persist~\citep{kim2025medical}, critically undermining the trust required for medical applications. 
Performance also varies significantly across languages~\citep{xie2024}, necessitating specialized models for contexts like Japan's healthcare system. 
A crucial, yet often overlooked, challenge is the observed performance degradation when complex reasoning is required to solve the task~\citep{chen2024huatuogpto1}, especially in multilingual contexts~\citep{xie2024}. 
Since clinicians often need to understand the model's rationale for verification and trust, this instability presents a major barrier to adoption.

Domain-specific adaptation is a common strategy for building LLMs for the medical domain. Continued Pretraining (CPT) on specialized corpora effectively infuses domain knowledge~\citep{gururangan2020dont, chen2023meditron, christophe2024beyond}, while instruction fine-tuning aligns models to specific tasks or styles~\citep{singhal2025medpalm, zhang2023huatuogpt}. 
While efforts exist for Japanese medical LLMs~\citep{sukeda202470b,sukeda2024}, they often use smaller models or focus on instruction following, potentially lacking deep knowledge integration on large models~\citep{gekhman2024does} or explicit optimization for reasoning stability.

To address the dual need for high accuracy and reliable reasoning explanations in the Japanese medical domain, we introduce Preferred-MedLLM-Qwen-72B. 
We propose a two-stage fine-tuning process applied to the Qwen2.5-72B foundation model~\citep{yang2024qwen2}. 
The first stage involves CPT using a comprehensive Japanese medical corpus, including the Japanese national licensing exam materials to embed deep domain knowledge. 
The second stage employs Reasoning Preference Optimization (RPO)~\citep{pang2024iterative}, a preference optimization technique extending Direct Preference Optimization (DPO)~\citep{Rafailov2023}, using a curated dataset comparing ground-truth and model-generated explanations. 
This RPO stage specifically aims to enhance the model's ability to generate stable, high-quality reasoning pathways while maintaining accuracy.

Our core contributions are:
\begin{itemize}
    \item We propose and evaluate a two-stage CPT+RPO fine-tuning approach designed to instill deep domain knowledge while specifically addressing the critical issue of performance instability when LLMs generate reasoning explanations in specialized domains.
    \item We demonstrate through evaluation on the Japanese Medical Licensing Exam benchmark (IgakuQA)~\citep{kasai2023igakuqa} that our resulting model, Preferred-MedLLM-Qwen-72B, achieves state-of-the-art accuracy (surpassing GPT-4o) and, crucially, maintains this high accuracy when required to provide explanations, validating the effectiveness of RPO for reasoning stabilization where CPT-only models falter.
    \item We release the Preferred-MedLLM-Qwen-72B model weights, providing a more reliable LLM for the Japanese medical domain and facilitating further research into trustworthy specialized LLMs.
\end{itemize}
This work underscores the importance of optimizing not just for accuracy but also for the reliability and transparency of the reasoning process, presenting a methodology to build more dependable LLMs for high-stakes, specialized domains and non-English languages. 

\section{Related Works}
The application of Large Language Models (LLMs) in medicine is rapidly evolving. Foundation models, such as OpenAI's GPT series~\citep{nori2023can}, o1 series~\citep{xie2024,brodeur2024} and Google's Gemini family~\citep{saab2024medgemini}, have demonstrated remarkable reasoning capabilities on various English medical benchmarks, often achieving high performance with minimal prompting~\citep{nori2024medprompto1}. However, significant challenges remain, including factual inaccuracies or hallucinations, and inconsistent performance across different languages and cultural contexts~\citep{xie2024}. These limitations hinder reliable deployment in high-stakes clinical settings and motivate the need for domain- and language-specific adaptations.

A primary strategy for adapting LLMs is further training on specialized data. Two common approaches are Continued Pretraining (CPT) and instruction fine-tuning. CPT involves extending the pretraining phase on large, domain-specific corpora. This approach, discussed conceptually by \citet{gururangan2020dont} and implemented in models like Meditron-70B~\citep{chen2023meditron} and others~\citep{christophe2024beyond}, is effective for infusing deep domain knowledge into the model, which is considered a more effective method for knowledge addition compared to instruction tuning~\citep{gekhman2024does}. Instruction fine-tuning, conversely, aligns models to follow specific instructions or interaction styles, as seen in models like MedPaLM for English medical QA~\citep{singhal2025medpalm}, HuatuoGPT for Chinese consultations~\citep{zhang2023huatuogpt}.

Work has also emerged specifically within the Japanese medical domain. For instance, \citet{sukeda2024} developed LLMs by applying instruction tuning to Japanese medical data. While valuable, such efforts have often utilized smaller base models (e.g., 7B parameters) or primarily focused on instruction following. This leaves a potential gap regarding the deep knowledge integration achievable via CPT on larger foundation models and the explicit optimization of reasoning processes, a critical step for ensuring reliability in real-world clinical and research applications.

Beyond fine-tuning model weights, prompt engineering techniques like Medprompt~\citep{nori2023medprompt} aim to elicit better medical performance from generalist models at inference time. However, the effectiveness and complexity of such strategies can vary~\citep{nori2024medprompto1}, potentially favoring methods that embed specialized knowledge and reasoning capabilities directly into the model parameters through training. Enhancing the inherent reasoning capabilities of LLMs is also an active research area, exploring techniques such as reinforcement learning~\citep{xie2024, guo2025deepseek}, reasoning with Reinforced Fine-Tuning (ReFT)~\citep{luong2024reft}, and preference optimization methods like Direct Preference Optimization (DPO)~\citep{Rafailov2023} and its variants~\citep{pang2024iterative}.

A critical gap, particularly relevant for clinical trust and adoption, is the performance degradation when models are required to generate step-by-step reasoning explanations. Our work addresses this specific challenge. We combine deep domain adaptation via CPT on a comprehensive Japanese medical corpus with Reasoning Preference Optimization (RPO)~\citep{pang2024iterative}, a DPO variant. RPO is specifically chosen here for its focus on not only learning preferences, but also on stabilizing the likelihood of the preferred reasoning path, aligning with our goal of generating reliable explanations alongside accurate answers. Unlike generalist models evaluated on medicine, or domain-specific models fine-tuned via CPT or instruction following, our approach explicitly targets the Japanese medical context and optimizes for both high accuracy and stable, high-quality reasoning explanations. We demonstrate that this combination achieves state-of-the-art performance on the complex Japanese Medical Licensing Exam (IgakuQA), surpassing strong proprietary models, and crucially, maintains this performance even when explicitly prompted for explanations, overcoming the instability observed in models without RPO.

\section{Methods}
Our approach involves a two-stage fine-tuning process applied to a powerful base LLM to create Preferred-MedLLM-Qwen-72B, specifically tailored for the Japanese medical domain with enhanced reasoning capabilities.

\subsection{Stage 1: Continued Pretraining (CPT) for Domain Adaptation}
\label{method:cpt}
We selected Qwen2.5-72B~\citep{yang2024qwen2} as the base model. This choice was based on its state-of-the-art performance on general benchmarks at the time of experimentation, its strong multilingual capabilities, which provide a solid base for Japanese language understanding.

The primary objective of the first stage was to infuse the base Qwen2.5-72B with comprehensive, up-to-date Japanese medical knowledge, addressing potential gaps not covered by its general pretraining corpus. In contrast to instruction tuning, which mainly adjusts response style based on existing knowledge~\citep{ouyang2022training}, CPT enables the model to acquire and integrate new domain-specific information~\citep{gekhman2024does}.

To achieve this, we utilized an original medical corpus, a collection of Japanese medical texts. This dataset includes explanations for past problems from the Japanese Medical Licensing Exam (JMLE) up to the year 2017. The use of JMLE materials only up to 2017 ensures no direct data leakage into the evaluation set, which covers 2018–2022~\citep{kasai2023igakuqa}.

Training a 72B parameter model presents significant computational challenges. We addressed this by employing QLoRA (Quantized Low-Rank Adaptation)~\citep{dettmers2023qlora}. Here, we combined 4-bit NormalFloat (NF4) quantization~\citep{dettmers2023case} with Low-Rank Adaptation (LoRA)~\citep{hu2021lora}. The CPT phase itself was conducted for two epochs. We employed the AdamW optimizer with cosine scheduling with warmup. This stage of training was carried out on an internal compute cluster, utilizing four NVIDIA A100 80GB GPUs.

\subsection{Stage 2: Reasoning Preference Optimization (RPO)}
Following CPT, the second stage focused on Reasoning Preference Optimization (RPO). The main goal here was to enhance the model's ability to generate not only accurate answers but also coherent and reliable reasoning explanations. This specifically aimed to address the performance instability observed in the CPT-only model when it was prompted to provide explanations alongside its answers.

To facilitate RPO, we curated a specialized preference dataset. This dataset was built using JMLE problems sourced up to 2017, ensuring distinction from the data used in IgakuQA. The curation process involved several steps. First, we prompted the CPT model (from Section~\ref{method:cpt}) using a three-shot template  (``Question: \{JMLE Question\}\textbackslash n\{Options\}\textbackslash n Explanation:\{Explanation for the question\}\textbackslash n Answer:\{Answer\}'']) to generate a response containing both a step-by-step explanation and a final conclusion answering the question.
Second, these generated responses were automatically categorized based on whether their final answer matched the ground truth correct answer for the respective JMLE problem. Responses with the correct final answer were labeled as ``Chosen Responses'' while those with incorrect final answers were labeled as ``Rejected Responses''. Third, we prepared ``Ground Truth Responses'', which consisted of the correct final answer paired with a verified explanation. It should be noted that these ground truth responses with verified explanations were also included in the data for continued pretraining.

With the preference data prepared, we employed Reasoning Preference Optimization (RPO)~\citep{pang2024iterative}, a variant of DPO~\citep{Rafailov2023}, implemented using the Hugging Face TRL library. RPO learns directly from preference pairs. We established a preference hierarchy for each JMLE problem: Ground Truth Response $>$ Chosen Response $>$ Rejected Response. This hierarchy guides the RPO loss computation, which consists of a weighted negative log-likelihood (NLL) loss on the chosen preferences together with the DPO loss, by creating preference pairs like (Ground Truth, Chosen), (Ground Truth, Rejected), and (Chosen, Rejected). This structure explicitly trains the model to favor responses demonstrating correctness and high-quality reasoning (represented by the ground truth preference) over those with correct answers but potentially weaker generated reasoning, while strongly penalizing incorrect answers. The underlying assumption is that the curated ground truth explanations embody a higher standard of reasoning compared to the model's intermediate generated explanations, even if the latter lead to a correct answer.

The RPO training phase was applied to the CPT model weights. For computational efficiency, we again utilized QLoRA. The model underwent training for one epoch over the entire preference dataset. We set the RPO alpha parameter, which weights the NLL loss relative to the DPO loss, to 10. This RPO stage was performed using two A100 GPUs.

\section{Results}
This section details the performance evaluation of Preferred-MedLLM-Qwen-72B. We first present its results on the primary benchmark, the Japanese National Medical Licensing Exam (IgakuQA), comparing it against baseline and proprietary models under different prompting conditions. We then provide an ablation study to dissect the contributions of the Continued Pretraining (CPT) and Reasoning Preference Optimization (RPO) stages. Finally, we assess the model's generalization capabilities on other relevant Japanese medical question-answering datasets.

\subsection{Performance on the IgakuQA Benchmark}
We evaluated our model using the IgakuQA benchmark~\citep{kasai2023igakuqa}, which comprises questions from the Japanese National Medical Licensing Exam spanning the years 2018 to 2022. Performance is reported as the average score achieved across these five years, calculated as the total points earned divided by the total possible points to account for varying question weights.

Evaluations were conducted under two distinct prompting scenarios. The first employed a standard 3-shot setting, where the prompt included three examples consisting of a question, options, and the correct answer, preceding the target question. The specific examples used were identical to those in the original IgakuQA implementation. As shown in Table~\ref{tab:igakuqa}, Preferred-MedLLM-Qwen-72B achieved a score of 0.868 in this standard setting. This performance surpasses that of the base Qwen2.5-72B (0.802) and the instruction-tuned Qwen2.5-72B-Instruct (0.802). It also slightly exceeds the score obtained by GPT-4o (0.866) and is considerably higher than GPT-4-Turbo (0.812) and the previously reported GPT-4 score (0.782)~\citep{kasai2023igakuqa}.

The second evaluation scenario, termed ``3-shot w/ explanation'', was designed specifically to assess performance stability when the model is explicitly required to generate its reasoning process. In this setting, the prompt instructed the model to provide a step-by-step explanation before delivering the final answer (similar in structure to the examples shown in Table~\ref{appendix_table}). Under this condition, Preferred-MedLLM-Qwen-72B maintained its high accuracy, achieving an identical score of 0.868. This stability contrasts with the behavior of the baseline Qwen2.5-72B model, whose performance decreased substantially from 0.802 in the standard setting to 0.710 when prompted for explanations. While GPT-4o demonstrated strong performance in this setting (0.881), our model outperformed the instruction-tuned Qwen2.5-72B-Instruct (0.822). These results underscore the effectiveness of our combined CPT and RPO methodology in yielding a model that not only achieves high accuracy but also preserves this accuracy when generating explanatory reasoning, a key desideratum for clinical utility.

\begin{table*}[tb]
\caption{Performance comparison on the IgakuQA benchmark. Scores represent the average scores over the 2018–2022 exams, normalized by the total score. Evaluations were conducted under standard 3-shot prompting and 3-shot prompting explicitly requiring explanations ('w/ explanation').\\ $^*$Scores for gpt-4 and gpt-3.5 were taken from \citet{kasai2023igakuqa}.}
\label{tab:igakuqa}
\vskip 0.15in
\begin{center}
\begin{small}
\begin{tabular}{lcc}
\toprule
\textbf{Model} & IgakuQA & IgakuQA\\
&(3-shot)& (3-shot w/ explanation)\\
\midrule
\textbf{Preferred-MedLLM-Qwen-72B} & \textbf{0.868} & 0.868 \\
gpt-4o & 0.866 & \textbf{0.881} \\
gpt-4-turbo & 0.812 & 0.814 \\
Qwen2.5-72B-Instruct & 0.802 & 0.822\\
Qwen2.5-72B & 0.802 & 0.710 \\
Llama3-Preferred-MedSwallow-70B & 0.795 & 0.744 \\
Llama-3.3-Swallow-70B-v0.4 & 0.787
 & 0.755\\
gpt-4$^*$ & 0.782 & - \\
gpt-4o-mini & 0.751 & 0.759 \\
Llama-3-Swallow-70B-v0.1 & 0.701 & 0.637 \\
sarashina2-70b & 0.561 & 0.549\\
gpt-3.5$^*$ & 0.550 & - \\
\bottomrule
\end{tabular}
\end{small}
\end{center}
\vskip -0.1in
\end{table*}

\subsection{Ablation Study: Impact of CPT and RPO}
To delineate the individual and combined contributions of CPT and RPO, we conducted an ablation study using the IgakuQA benchmark. The results are presented in Table~\ref{tab:ablation}. We also included a comparison using standard DPO instead of RPO.

The baseline Qwen2.5-72B model achieved scores of 0.802 (3-shot) and 0.710 (3-shot w/ explanation). Applying only CPT (Stage 1) significantly improved performance in the standard setting to 0.867, confirming the effectiveness of CPT for domain knowledge infusion. However, this CPT-only model exhibited a performance drop to 0.834 when explanations were required, highlighting the reasoning instability issue motivating our work. Conversely, applying RPO directly to the base Qwen2.5-72B model resulted in scores of 0.838 (3-shot) and 0.807 (3-shot w/ explanation). This indicates that RPO alone can enhance both accuracy and reasoning stability compared to the baseline, likely by better aligning the model's inherent reasoning pathways, although it lacks the deep domain knowledge provided by CPT.

Combining CPT with standard DPO yielded scores of 0.868 (3-shot) and 0.848 (3-shot w/ explanation). While achieving high accuracy comparable to our final model in the standard setting, this configuration still showed a noticeable decrease in performance when generating explanations.

Finally, our proposed two-stage approach, combining CPT with RPO (Preferred-MedLLM-Qwen-72B), achieved a score of 0.868 in the standard 3-shot setting and, crucially, maintained this exact score of 0.868 in the 3-shot setting requiring explanations. This demonstrates perfect stability under explanation generation for this benchmark. These ablation results provide strong evidence for the synergy between CPT and RPO. CPT is essential for incorporating domain-specific knowledge, while RPO effectively refines the model's reasoning generation, ensuring consistent accuracy even when explanations are explicitly prompted. Furthermore, the comparison between CPT+DPO and CPT+RPO suggests that RPO, potentially due to its inclusion of an NLL loss component alongside the DPO loss, offers superior stabilization for the reasoning process in this context.

\begin{table*}[t]
\caption{Ablation study on IgakuQA showing the impact of CPT, RPO, and DPO applied to Qwen2.5-72B. Scores represent average accuracy under standard 3-shot and 3-shot with explanation ('w/ explanation') settings.}
\label{tab:ablation}
\vskip 0.15in
\begin{center}
\begin{small}
\begin{tabular}{lcc}
\hline
\textbf{Model} & IgakuQA & IgakuQA\\
&(3-shot)&(3-shot w/ explanation)\\
\midrule
Preferred-MedLLM-Qwen-72B (CPT+RPO) & \textbf{0.868} & \textbf{0.868} \\
Qwen2.5-72B + CPT + DPO & \textbf{0.868} & 0.848 \\
Qwen2.5-72B + CPT & 0.867 & 0.834 \\
Qwen2.5-72B + RPO & 0.838 & 0.807 \\
Qwen2.5-72B & 0.802 & 0.710 \\
\bottomrule
\end{tabular}
\end{small}
\end{center}
\vskip -0.1in
\end{table*}

\subsection{Performance on Other Japanese Medical Benchmarks}
To evaluate the generalizability of the improvements imparted by our fine-tuning methodology, we assessed Preferred-MedLLM-Qwen-72B on a selection of other Japanese medical question-answering benchmarks. These evaluations were performed in a zero-shot setting to probe the model's inherent capabilities without task-specific examples. The benchmarks included Japanese translations of MedQA~\citep{jin2020disease}, MedMCQA~\citep{pal2022medmcqa}, PubMedQA~\citep{jin-etal-2019-pubmedqa} (using translations from \citet{jiang2024jmedbench}), and the medicine related subset (``anatomy'', ``clinical\_knowledge'', ``college\_medicine'', ``medical\_genetics'', ``professional\_medicine'') of MMMLU~\citep{mmmlu, hendrycks2020measuring}.

The results, summarized in Table~\ref{tab:other_benchmarks}, indicate that Preferred-MedLLM-Qwen-72B generally outperforms both the base Qwen2.5-72B and the Qwen2.5-72B-Instruct models across these diverse tasks. Our model achieved the highest scores on the Japanese medical related tasks of MMMLU, and the Japanese translated versions of MedQA, and MedMCQA, resulting in the highest average score (0.716) among the compared models. Furthermore, comparing our final Preferred-MedLLM-Qwen-72B model (CPT+RPO) to the model after only the CPT stage (Qwen2.5-72B + CPT in Table~\ref{tab:other_benchmarks}), we observe that the addition of RPO yields further small but consistent improvements across most of these diverse benchmarks, increasing the average score slightly from 0.710 to 0.716. This consistent performance advantage suggests that the benefits of the combined CPT and RPO stages, including the potential positive impact of RPO on generalizability, extend effectively beyond the specific format of the Japanese Medical Licensing Exam to broader medical question-answering scenarios in Japanese.

\begin{table*}[tb]
\caption{Zero-shot performance on various Japanese medical QA benchmarks. Scores represent accuracy on Japanese translations$^\dagger$ of MedQA, MedMCQA, PubMedQA, and relevant subsets (``anatomy'', ``clinical\_knowledge'', ``college\_medicine'', ``medical\_genetics'', ``professional\_medicine'') of MMMLU.\\ $^\dagger$Benchmark translations sourced from \citet{jiang2024jmedbench}.}
\label{tab:other_benchmarks}
\vskip 0.15in
\begin{center}
\begin{small}
\begin{tabular}{lccccc}
\toprule
\textbf{Model (0-shot)} & MMMLU (med, jp) & MedQA (jp) & MedMCQA (jp) & PubMedQA (jp) & Average\\
\midrule
Preferred-MedLLM-Qwen-72B & \textbf{0.800} & \textbf{0.684} & \textbf{0.602} & \textbf{0.779} & \textbf{0.716}\\
Qwen2.5-72B + CPT & 0.799 & 0.669 & 0.598 & 0.775 & 0.710 \\
Qwen2.5-72B & 0.797 & 0.652 & 0.591 & \textbf{0.779} & 0.705 \\
Qwen2.5-72B-Instruct & 0.795 & 0.626 & 0.601 & 0.714 & 0.684\\
\bottomrule
\end{tabular}
\end{small}
\end{center}
\vskip -0.1in
\end{table*}

\section{Conclusion and Discussions}
In this work, we introduced Preferred-MedLLM-Qwen-72B, a large language model specifically fine-tuned for the Japanese medical domain.
Our primary goal was to develop an LLM exhibiting not only high accuracy but also reliable performance when generating reasoning explanations crucial for clinical trust.
We proposed a two-stage fine-tuning process combining Continued Pretraining (CPT) for deep domain knowledge infusion and Reasoning Preference Optimization (RPO) for enhancing reasoning alignment and stability.
Our core contribution lies in demonstrating that applying RPO subsequent to CPT effectively mitigates the reasoning instability—a performance degradation observed when explicit explanations are required—that can manifest in models trained with CPT alone or even CPT combined with standard DPO.

Our empirical results validate this approach. Preferred-MedLLM-Qwen-72B achieves state-of-the-art performance on the challenging Japanese Medical Licensing Exam benchmark (IgakuQA), surpassing its base model and competitive proprietary models like GPT-4o in standard evaluations.
More significantly, ablation studies confirmed that while CPT is vital for knowledge and accuracy gains, the RPO stage is crucial for maintaining this high performance when the model is prompted to generate step-by-step explanations.
This stability under explanation is a key differentiator from baseline and CPT-only models. 
Furthermore, the model demonstrated strong performance across other Japanese medical QA benchmarks, suggesting the benefits of our CPT+RPO methodology generalize beyond the specific JMLE task format.
This work presents a promising methodology for building more reliable and transparent AI systems by strategically optimizing for both accuracy and stable reasoning generation.

While these findings are encouraging, we acknowledge several limitations and identify avenues for future investigation.
First, our evaluation primarily focused on multiple-choice question-answering benchmarks. 
Assessing performance on a broader spectrum of clinical tasks, such as medical report generation~\citep{kanithi2024medic}, summarization~\citep{van2024adapted}, and dialogue systems~\citep{johri2025evaluation}, is essential to fully understand the model's capabilities.
Second, the preference data used for RPO was generated semi-automatically based on ground truth answers and model outputs.
Future work could explore incorporating expert human feedback into the preference framework, which might further enhance reasoning quality, although scalability must be considered.
Third, we did not investigate combining RPO with instruction tuning; exploring this combination could reveal complementary benefits for reasoning stabilization and overall performance. 
Fourth, the significant computational resources required to train and deploy 72B parameter models pose a practical challenge; future research should explore applying our CPT+RPO methodology to smaller, more efficient architectures or investigate model distillation techniques. 
Finally, rigorous real-world clinical validation remains a critical next step to evaluate the practical utility, safety, and trustworthiness of the model and its explanations within actual healthcare settings.

In conclusion, Preferred-MedLLM-Qwen-72B represents a significant step towards developing specialized LLMs suitable for demanding domains like Japanese medicine.
By strategically combining CPT for knowledge and RPO for reasoning alignment, we have created a model that achieves high accuracy while crucially maintaining performance stability during explanation generation.
This highlights an effective pathway for building more reliable and transparent AI systems, paving the way for their responsible integration into clinical practice and other high-stakes applications.

\section*{Software and Data}
The trained model can be accessed at \url{https://huggingface.co/pfnet/Preferred-MedLLM-Qwen-72B}.

\section*{Acknowledgements}
We would like to thank the Preferred Networks cluster team members for the infrastructure support.

\bibliography{ref}

\begin{thebibliography}{41}
\providecommand{\natexlab}[1]{#1}
\providecommand{\url}[1]{\texttt{#1}}
\expandafter\ifx\csname urlstyle\endcsname\relax
  \providecommand{\doi}[1]{doi: #1}\else
  \providecommand{\doi}{doi: \begingroup \urlstyle{rm}\Url}\fi

\bibitem[Brodeur et~al.(2024)Brodeur, Buckley, Kanjee, Goh, Ling, Jain, Cabral, Abdulnour, Haimovich, Freed, Olson, Morgan, Hom, Gallo, Horvitz, Chen, Manrai, and Rodman]{brodeur2024}
Peter~G. Brodeur, Thomas~A. Buckley, Zahir Kanjee, Ethan Goh, Evelyn~Bin Ling, Priyank Jain, Stephanie Cabral, Raja-Elie Abdulnour, Adrian Haimovich, Jason~A. Freed, Andrew Olson, Daniel~J. Morgan, Jason Hom, Robert Gallo, Eric Horvitz, Jonathan Chen, Arjun~K. Manrai, and Adam Rodman.
\newblock Superhuman performance of a large language model on the reasoning tasks of a physician.
\newblock \emph{arXiv preprint arXiv:2412.10849}, 2024.
\newblock URL \url{https://arxiv.org/abs/2412.10849}.

\bibitem[Chen et~al.(2024)Chen, Cai, Ji, Wang, Liu, Wang, Hou, and Wang]{chen2024huatuogpto1}
Junying Chen, Zhenyang Cai, Ke~Ji, Xidong Wang, Wanlong Liu, Rongsheng Wang, Jianye Hou, and Benyou Wang.
\newblock Huatuogpt-o1, towards medical complex reasoning with llms.
\newblock \emph{arXiv preprint arXiv:2412.18925}, 2024.
\newblock URL \url{https://arxiv.org/abs/2412.18925}.

\bibitem[Chen et~al.(2023)Chen, Cano, Romanou, Bonnet, Matoba, Salvi, Pagliardini, Fan, Köpf, Mohtashami, Sallinen, Sakhaeirad, Swamy, Krawczuk, Bayazit, Marmet, Montariol, Hartley, Jaggi, and Bosselut]{chen2023meditron}
Zeming Chen, Alejandro~Hernández Cano, Angelika Romanou, Antoine Bonnet, Kyle Matoba, Francesco Salvi, Matteo Pagliardini, Simin Fan, Andreas Köpf, Amirkeivan Mohtashami, Alexandre Sallinen, Alireza Sakhaeirad, Vinitra Swamy, Igor Krawczuk, Deniz Bayazit, Axel Marmet, Syrielle Montariol, Mary-Anne Hartley, Martin Jaggi, and Antoine Bosselut.
\newblock Meditron-70b: Scaling medical pretraining for large language models.
\newblock \emph{arXiv preprint arXiv:2311.16079}, 2023.
\newblock URL \url{https://arxiv.org/abs/2311.16079}.

\bibitem[Christophe et~al.(2024)Christophe, Raha, Maslenkova, Salman, Kanithi, Pimentel, and Khan]{christophe2024beyond}
Clement Christophe, Tathagata Raha, Svetlana Maslenkova, Muhammad~Umar Salman, Praveenkumar Kanithi, Marco~AF Pimentel, and Shadab Khan.
\newblock Beyond fine-tuning: Unleashing the potential of continuous pretraining for clinical {LLM}s.
\newblock In Yaser Al-Onaizan, Mohit Bansal, and Yun-Nung Chen, editors, \emph{Findings of the Association for Computational Linguistics: EMNLP 2024}, pages 10549--10561, Miami, Florida, USA, November 2024. Association for Computational Linguistics.
\newblock \doi{10.18653/v1/2024.findings-emnlp.618}.
\newblock URL \url{https://aclanthology.org/2024.findings-emnlp.618/}.

\bibitem[{Cosentino} et~al.(2024){Cosentino}, {Belyaeva}, {Liu}, {Furlotte}, {Yang}, {Lee}, {Schenck}, {Patel}, {Cui}, {Schneider}, {Bryant}, {Gomes}, {Jiang}, {Lee}, {Liu}, {Perez}, {Rogers}, {Speed}, {Tailor}, {Walker}, {Yu}, {Althoff}, {Heneghan}, {Hernandez}, {Malhotra}, {Stern}, {Matias}, {Corrado}, {Patel}, {Shetty}, {Zhan}, {Prabhakara}, {McDuff}, and {McLean}]{cosentino2024towards}
Justin {Cosentino}, Anastasiya {Belyaeva}, Xin {Liu}, Nicholas~A. {Furlotte}, Zhun {Yang}, Chace {Lee}, Erik {Schenck}, Yojan {Patel}, Jian {Cui}, Logan~Douglas {Schneider}, Robby {Bryant}, Ryan~G. {Gomes}, Allen {Jiang}, Roy {Lee}, Yun {Liu}, Javier {Perez}, Jameson~K. {Rogers}, Cathy {Speed}, Shyam {Tailor}, Megan {Walker}, Jeffrey {Yu}, Tim {Althoff}, Conor {Heneghan}, John {Hernandez}, Mark {Malhotra}, Leor {Stern}, Yossi {Matias}, Greg~S. {Corrado}, Shwetak {Patel}, Shravya {Shetty}, Jiening {Zhan}, Shruthi {Prabhakara}, Daniel {McDuff}, and Cory~Y. {McLean}.
\newblock {Towards a Personal Health Large Language Model}.
\newblock \emph{arXiv e-prints}, art. arXiv:2406.06474, June 2024.
\newblock \doi{10.48550/arXiv.2406.06474}.

\bibitem[{DeepSeek-AI} et~al.(2025){DeepSeek-AI}, {Guo}, {Yang}, {Zhang}, {Song}, {Zhang}, {Xu}, {Zhu}, {Ma}, {Wang}, {Bi}, {Zhang}, {Yu}, {Wu}, {Wu}, {Gou}, {Shao}, {Li}, {Gao}, {Liu}, {Xue}, {Wang}, {Wu}, {Feng}, {Lu}, {Zhao}, {Deng}, {Zhang}, {Ruan}, {Dai}, {Chen}, {Ji}, {Li}, {Lin}, {Dai}, {Luo}, {Hao}, {Chen}, {Li}, {Zhang}, {Bao}, {Xu}, {Wang}, {Ding}, {Xin}, {Gao}, {Qu}, {Li}, {Guo}, {Li}, {Wang}, {Chen}, {Yuan}, {Qiu}, {Li}, {Cai}, {Ni}, {Liang}, {Chen}, {Dong}, {Hu}, {Gao}, {Guan}, {Huang}, {Yu}, {Wang}, {Zhang}, {Zhao}, {Wang}, {Zhang}, {Xu}, {Xia}, {Zhang}, {Zhang}, {Tang}, {Li}, {Wang}, {Li}, {Tian}, {Huang}, {Zhang}, {Wang}, {Chen}, {Du}, {Ge}, {Zhang}, {Pan}, {Wang}, {Chen}, {Jin}, {Chen}, {Lu}, {Zhou}, {Chen}, {Ye}, {Wang}, {Yu}, {Zhou}, {Pan}, {Li}, {Zhou}, {Wu}, {Ye}, {Yun}, {Pei}, {Sun}, {Wang}, {Zeng}, {Zhao}, {Liu}, {Liang}, {Gao}, {Yu}, {Zhang}, {Xiao}, {An}, {Liu}, {Wang}, {Chen}, {Nie}, {Cheng}, {Liu}, {Xie}, {Liu}, {Yang}, {Li}, {Su}, {Lin}, {Li}, {Jin}, {Shen}, {Chen}, {Sun}, {Wang},
  {Song}, {Zhou}, {Wang}, {Shan}, {Li}, {Wang}, {Wei}, {Zhang}, {Xu}, {Li}, {Zhao}, {Sun}, {Wang}, {Yu}, {Zhang}, {Shi}, {Xiong}, {He}, {Piao}, {Wang}, {Tan}, {Ma}, {Liu}, {Guo}, {Ou}, {Wang}, {Gong}, {Zou}, {He}, {Xiong}, {Luo}, {You}, {Liu}, {Zhou}, {Zhu}, {Xu}, {Huang}, {Li}, {Zheng}, {Zhu}, {Ma}, {Tang}, {Zha}, {Yan}, {Ren}, {Ren}, {Sha}, {Fu}, {Xu}, {Xie}, {Zhang}, {Hao}, {Ma}, {Yan}, {Wu}, {Gu}, {Zhu}, {Liu}, {Li}, {Xie}, {Song}, {Pan}, {Huang}, {Xu}, {Zhang}, and {Zhang}]{guo2025deepseek}
{DeepSeek-AI}, Daya {Guo}, Dejian {Yang}, Haowei {Zhang}, Junxiao {Song}, Ruoyu {Zhang}, Runxin {Xu}, Qihao {Zhu}, Shirong {Ma}, Peiyi {Wang}, Xiao {Bi}, Xiaokang {Zhang}, Xingkai {Yu}, Yu~{Wu}, Z.~F. {Wu}, Zhibin {Gou}, Zhihong {Shao}, Zhuoshu {Li}, Ziyi {Gao}, Aixin {Liu}, Bing {Xue}, Bingxuan {Wang}, Bochao {Wu}, Bei {Feng}, Chengda {Lu}, Chenggang {Zhao}, Chengqi {Deng}, Chenyu {Zhang}, Chong {Ruan}, Damai {Dai}, Deli {Chen}, Dongjie {Ji}, Erhang {Li}, Fangyun {Lin}, Fucong {Dai}, Fuli {Luo}, Guangbo {Hao}, Guanting {Chen}, Guowei {Li}, H.~{Zhang}, Han {Bao}, Hanwei {Xu}, Haocheng {Wang}, Honghui {Ding}, Huajian {Xin}, Huazuo {Gao}, Hui {Qu}, Hui {Li}, Jianzhong {Guo}, Jiashi {Li}, Jiawei {Wang}, Jingchang {Chen}, Jingyang {Yuan}, Junjie {Qiu}, Junlong {Li}, J.~L. {Cai}, Jiaqi {Ni}, Jian {Liang}, Jin {Chen}, Kai {Dong}, Kai {Hu}, Kaige {Gao}, Kang {Guan}, Kexin {Huang}, Kuai {Yu}, Lean {Wang}, Lecong {Zhang}, Liang {Zhao}, Litong {Wang}, Liyue {Zhang}, Lei {Xu}, Leyi {Xia}, Mingchuan {Zhang}, Minghua
  {Zhang}, Minghui {Tang}, Meng {Li}, Miaojun {Wang}, Mingming {Li}, Ning {Tian}, Panpan {Huang}, Peng {Zhang}, Qiancheng {Wang}, Qinyu {Chen}, Qiushi {Du}, Ruiqi {Ge}, Ruisong {Zhang}, Ruizhe {Pan}, Runji {Wang}, R.~J. {Chen}, R.~L. {Jin}, Ruyi {Chen}, Shanghao {Lu}, Shangyan {Zhou}, Shanhuang {Chen}, Shengfeng {Ye}, Shiyu {Wang}, Shuiping {Yu}, Shunfeng {Zhou}, Shuting {Pan}, S.~S. {Li}, Shuang {Zhou}, Shaoqing {Wu}, Shengfeng {Ye}, Tao {Yun}, Tian {Pei}, Tianyu {Sun}, T.~{Wang}, Wangding {Zeng}, Wanjia {Zhao}, Wen {Liu}, Wenfeng {Liang}, Wenjun {Gao}, Wenqin {Yu}, Wentao {Zhang}, W.~L. {Xiao}, Wei {An}, Xiaodong {Liu}, Xiaohan {Wang}, Xiaokang {Chen}, Xiaotao {Nie}, Xin {Cheng}, Xin {Liu}, Xin {Xie}, Xingchao {Liu}, Xinyu {Yang}, Xinyuan {Li}, Xuecheng {Su}, Xuheng {Lin}, X.~Q. {Li}, Xiangyue {Jin}, Xiaojin {Shen}, Xiaosha {Chen}, Xiaowen {Sun}, Xiaoxiang {Wang}, Xinnan {Song}, Xinyi {Zhou}, Xianzu {Wang}, Xinxia {Shan}, Y.~K. {Li}, Y.~Q. {Wang}, Y.~X. {Wei}, Yang {Zhang}, Yanhong {Xu}, Yao {Li}, Yao
  {Zhao}, Yaofeng {Sun}, Yaohui {Wang}, Yi~{Yu}, Yichao {Zhang}, Yifan {Shi}, Yiliang {Xiong}, Ying {He}, Yishi {Piao}, Yisong {Wang}, Yixuan {Tan}, Yiyang {Ma}, Yiyuan {Liu}, Yongqiang {Guo}, Yuan {Ou}, Yuduan {Wang}, Yue {Gong}, Yuheng {Zou}, Yujia {He}, Yunfan {Xiong}, Yuxiang {Luo}, Yuxiang {You}, Yuxuan {Liu}, Yuyang {Zhou}, Y.~X. {Zhu}, Yanhong {Xu}, Yanping {Huang}, Yaohui {Li}, Yi~{Zheng}, Yuchen {Zhu}, Yunxian {Ma}, Ying {Tang}, Yukun {Zha}, Yuting {Yan}, Z.~Z. {Ren}, Zehui {Ren}, Zhangli {Sha}, Zhe {Fu}, Zhean {Xu}, Zhenda {Xie}, Zhengyan {Zhang}, Zhewen {Hao}, Zhicheng {Ma}, Zhigang {Yan}, Zhiyu {Wu}, Zihui {Gu}, Zijia {Zhu}, Zijun {Liu}, Zilin {Li}, Ziwei {Xie}, Ziyang {Song}, Zizheng {Pan}, Zhen {Huang}, Zhipeng {Xu}, Zhongyu {Zhang}, and Zhen {Zhang}.
\newblock {DeepSeek-R1: Incentivizing Reasoning Capability in LLMs via Reinforcement Learning}.
\newblock \emph{arXiv e-prints}, art. arXiv:2501.12948, January 2025.
\newblock \doi{10.48550/arXiv.2501.12948}.

\bibitem[{Dettmers} and {Zettlemoyer}(2022)]{dettmers2023case}
Tim {Dettmers} and Luke {Zettlemoyer}.
\newblock {The case for 4-bit precision: k-bit Inference Scaling Laws}.
\newblock \emph{arXiv e-prints}, art. arXiv:2212.09720, December 2022.
\newblock \doi{10.48550/arXiv.2212.09720}.

\bibitem[Dettmers et~al.(2023)Dettmers, Pagnoni, Holtzman, and Zettlemoyer]{dettmers2023qlora}
Tim Dettmers, Artidoro Pagnoni, Ari Holtzman, and Luke Zettlemoyer.
\newblock Qlora: Efficient finetuning of quantized llms.
\newblock \emph{Advances in neural information processing systems}, 36:\penalty0 10088--10115, 2023.

\bibitem[Gekhman et~al.(2024)Gekhman, Yona, Aharoni, Eyal, Feder, Reichart, and Herzig]{gekhman2024does}
Zorik Gekhman, Gal Yona, Roee Aharoni, Matan Eyal, Amir Feder, Roi Reichart, and Jonathan Herzig.
\newblock Does fine-tuning llms on new knowledge encourage hallucinations?
\newblock \emph{arXiv preprint arXiv:2405.05904}, 2024.

\bibitem[{Grattafiori} et~al.(2024){Grattafiori}, {Dubey}, {Jauhri}, {Pandey}, {Kadian}, {Al-Dahle}, {Letman}, {Mathur}, {Schelten}, {Vaughan}, {Yang}, {Fan}, {Goyal}, {Hartshorn}, {Yang}, {Mitra}, {Sravankumar}, {Korenev}, {Hinsvark}, {Rao}, {Zhang}, {Rodriguez}, {Gregerson}, {Spataru}, {Roziere}, {Biron}, {Tang}, {Chern}, {Caucheteux}, {Nayak}, {Bi}, {Marra}, {McConnell}, {Keller}, {Touret}, {Wu}, {Wong}, {Canton Ferrer}, {Nikolaidis}, {Allonsius}, {Song}, {Pintz}, {Livshits}, {Wyatt}, {Esiobu}, {Choudhary}, {Mahajan}, {Garcia-Olano}, {Perino}, {Hupkes}, {Lakomkin}, {AlBadawy}, {Lobanova}, {Dinan}, {Smith}, {Radenovic}, {Guzm{\'a}n}, {Zhang}, {Synnaeve}, {Lee}, {Anderson}, {Thattai}, {Nail}, {Mialon}, {Pang}, {Cucurell}, {Nguyen}, {Korevaar}, {Xu}, {Touvron}, {Zarov}, {Arrieta Ibarra}, {Kloumann}, {Misra}, {Evtimov}, {Zhang}, {Copet}, {Lee}, {Geffert}, {Vranes}, {Park}, {Mahadeokar}, {Shah}, {van der Linde}, {Billock}, {Hong}, {Lee}, {Fu}, {Chi}, {Huang}, {Liu}, {Wang}, {Yu}, {Bitton}, {Spisak}, {Park},
  {Rocca}, {Johnstun}, {Saxe}, {Jia}, {Vasuden Alwala}, {Prasad}, {Upasani}, {Plawiak}, {Li}, {Heafield}, {Stone}, {El-Arini}, {Iyer}, {Malik}, {Chiu}, {Bhalla}, {Lakhotia}, {Rantala-Yeary}, {van der Maaten}, {Chen}, {Tan}, {Jenkins}, {Martin}, {Madaan}, {Malo}, {Blecher}, {Landzaat}, {de Oliveira}, {Muzzi}, {Pasupuleti}, {Singh}, {Paluri}, {Kardas}, {Tsimpoukelli}, {Oldham}, {Rita}, {Pavlova}, {Kambadur}, {Lewis}, {Si}, {Singh}, {Hassan}, {Goyal}, {Torabi}, {Bashlykov}, {Bogoychev}, {Chatterji}, {Zhang}, {Duchenne}, {{\c{C}}elebi}, {Alrassy}, {Zhang}, {Li}, {Vasic}, {Weng}, {Bhargava}, {Dubal}, {Krishnan}, {Singh Koura}, {Xu}, {He}, {Dong}, {Srinivasan}, {Ganapathy}, {Calderer}, {Silveira Cabral}, {Stojnic}, {Raileanu}, {Maheswari}, {Girdhar}, {Patel}, {Sauvestre}, {Polidoro}, {Sumbaly}, {Taylor}, {Silva}, {Hou}, {Wang}, {Hosseini}, {Chennabasappa}, {Singh}, {Bell}, {Kim}, {Edunov}, {Nie}, {Narang}, {Raparthy}, {Shen}, {Wan}, {Bhosale}, {Zhang}, {Vandenhende}, {Batra}, {Whitman}, {Sootla}, {Collot},
  {Gururangan}, {Borodinsky}, {Herman}, {Fowler}, {Sheasha}, {Georgiou}, {Scialom}, and {Speckbacher}]{grattafiori2024llama}
Aaron {Grattafiori}, Abhimanyu {Dubey}, Abhinav {Jauhri}, Abhinav {Pandey}, Abhishek {Kadian}, Ahmad {Al-Dahle}, Aiesha {Letman}, Akhil {Mathur}, Alan {Schelten}, Alex {Vaughan}, Amy {Yang}, Angela {Fan}, Anirudh {Goyal}, Anthony {Hartshorn}, Aobo {Yang}, Archi {Mitra}, Archie {Sravankumar}, Artem {Korenev}, Arthur {Hinsvark}, Arun {Rao}, Aston {Zhang}, Aurelien {Rodriguez}, Austen {Gregerson}, Ava {Spataru}, Baptiste {Roziere}, Bethany {Biron}, Binh {Tang}, Bobbie {Chern}, Charlotte {Caucheteux}, Chaya {Nayak}, Chloe {Bi}, Chris {Marra}, Chris {McConnell}, Christian {Keller}, Christophe {Touret}, Chunyang {Wu}, Corinne {Wong}, Cristian {Canton Ferrer}, Cyrus {Nikolaidis}, Damien {Allonsius}, Daniel {Song}, Danielle {Pintz}, Danny {Livshits}, Danny {Wyatt}, David {Esiobu}, Dhruv {Choudhary}, Dhruv {Mahajan}, Diego {Garcia-Olano}, Diego {Perino}, Dieuwke {Hupkes}, Egor {Lakomkin}, Ehab {AlBadawy}, Elina {Lobanova}, Emily {Dinan}, Eric~Michael {Smith}, Filip {Radenovic}, Francisco {Guzm{\'a}n}, Frank {Zhang},
  Gabriel {Synnaeve}, Gabrielle {Lee}, Georgia~Lewis {Anderson}, Govind {Thattai}, Graeme {Nail}, Gregoire {Mialon}, Guan {Pang}, Guillem {Cucurell}, Hailey {Nguyen}, Hannah {Korevaar}, Hu~{Xu}, Hugo {Touvron}, Iliyan {Zarov}, Imanol {Arrieta Ibarra}, Isabel {Kloumann}, Ishan {Misra}, Ivan {Evtimov}, Jack {Zhang}, Jade {Copet}, Jaewon {Lee}, Jan {Geffert}, Jana {Vranes}, Jason {Park}, Jay {Mahadeokar}, Jeet {Shah}, Jelmer {van der Linde}, Jennifer {Billock}, Jenny {Hong}, Jenya {Lee}, Jeremy {Fu}, Jianfeng {Chi}, Jianyu {Huang}, Jiawen {Liu}, Jie {Wang}, Jiecao {Yu}, Joanna {Bitton}, Joe {Spisak}, Jongsoo {Park}, Joseph {Rocca}, Joshua {Johnstun}, Joshua {Saxe}, Junteng {Jia}, Kalyan {Vasuden Alwala}, Karthik {Prasad}, Kartikeya {Upasani}, Kate {Plawiak}, Ke~{Li}, Kenneth {Heafield}, Kevin {Stone}, Khalid {El-Arini}, Krithika {Iyer}, Kshitiz {Malik}, Kuenley {Chiu}, Kunal {Bhalla}, Kushal {Lakhotia}, Lauren {Rantala-Yeary}, Laurens {van der Maaten}, Lawrence {Chen}, Liang {Tan}, Liz {Jenkins}, Louis {Martin},
  Lovish {Madaan}, Lubo {Malo}, Lukas {Blecher}, Lukas {Landzaat}, Luke {de Oliveira}, Madeline {Muzzi}, Mahesh {Pasupuleti}, Mannat {Singh}, Manohar {Paluri}, Marcin {Kardas}, Maria {Tsimpoukelli}, Mathew {Oldham}, Mathieu {Rita}, Maya {Pavlova}, Melanie {Kambadur}, Mike {Lewis}, Min {Si}, Mitesh~Kumar {Singh}, Mona {Hassan}, Naman {Goyal}, Narjes {Torabi}, Nikolay {Bashlykov}, Nikolay {Bogoychev}, Niladri {Chatterji}, Ning {Zhang}, Olivier {Duchenne}, Onur {{\c{C}}elebi}, Patrick {Alrassy}, Pengchuan {Zhang}, Pengwei {Li}, Petar {Vasic}, Peter {Weng}, Prajjwal {Bhargava}, Pratik {Dubal}, Praveen {Krishnan}, Punit {Singh Koura}, Puxin {Xu}, Qing {He}, Qingxiao {Dong}, Ragavan {Srinivasan}, Raj {Ganapathy}, Ramon {Calderer}, Ricardo {Silveira Cabral}, Robert {Stojnic}, Roberta {Raileanu}, Rohan {Maheswari}, Rohit {Girdhar}, Rohit {Patel}, Romain {Sauvestre}, Ronnie {Polidoro}, Roshan {Sumbaly}, Ross {Taylor}, Ruan {Silva}, Rui {Hou}, Rui {Wang}, Saghar {Hosseini}, Sahana {Chennabasappa}, Sanjay {Singh}, Sean
  {Bell}, Seohyun~Sonia {Kim}, Sergey {Edunov}, Shaoliang {Nie}, Sharan {Narang}, Sharath {Raparthy}, Sheng {Shen}, Shengye {Wan}, Shruti {Bhosale}, Shun {Zhang}, Simon {Vandenhende}, Soumya {Batra}, Spencer {Whitman}, Sten {Sootla}, Stephane {Collot}, Suchin {Gururangan}, Sydney {Borodinsky}, Tamar {Herman}, Tara {Fowler}, Tarek {Sheasha}, Thomas {Georgiou}, Thomas {Scialom}, and Tobias {Speckbacher}.
\newblock {The Llama 3 Herd of Models}.
\newblock \emph{arXiv e-prints}, art. arXiv:2407.21783, July 2024.
\newblock \doi{10.48550/arXiv.2407.21783}.

\bibitem[Gururangan et~al.(2020)Gururangan, Marasovi{\'c}, Swayamdipta, Lo, Beltagy, Downey, and Smith]{gururangan2020dont}
Suchin Gururangan, Ana Marasovi{\'c}, Swabha Swayamdipta, Kyle Lo, Iz~Beltagy, Doug Downey, and Noah~A. Smith.
\newblock Don`t stop pretraining: Adapt language models to domains and tasks.
\newblock In Dan Jurafsky, Joyce Chai, Natalie Schluter, and Joel Tetreault, editors, \emph{Proceedings of the 58th Annual Meeting of the Association for Computational Linguistics}, pages 8342--8360, Online, July 2020. Association for Computational Linguistics.
\newblock \doi{10.18653/v1/2020.acl-main.740}.
\newblock URL \url{https://aclanthology.org/2020.acl-main.740/}.

\bibitem[{Hendrycks} et~al.(2020){Hendrycks}, {Burns}, {Basart}, {Zou}, {Mazeika}, {Song}, and {Steinhardt}]{hendrycks2020measuring}
Dan {Hendrycks}, Collin {Burns}, Steven {Basart}, Andy {Zou}, Mantas {Mazeika}, Dawn {Song}, and Jacob {Steinhardt}.
\newblock {Measuring Massive Multitask Language Understanding}.
\newblock \emph{arXiv e-prints}, art. arXiv:2009.03300, September 2020.
\newblock \doi{10.48550/arXiv.2009.03300}.

\bibitem[Hu et~al.(2021)Hu, Shen, Wallis, Allen-Zhu, Li, Wang, Wang, and Chen]{hu2021lora}
Edward~J Hu, Yelong Shen, Phillip Wallis, Zeyuan Allen-Zhu, Yuanzhi Li, Shean Wang, Lu~Wang, and Weizhu Chen.
\newblock Lora: Low-rank adaptation of large language models. arxiv 2021.
\newblock \emph{arXiv preprint arXiv:2106.09685}, 2021.

\bibitem[{Jiang} et~al.(2024){Jiang}, {Huang}, and {Aizawa}]{jiang2024jmedbench}
Junfeng {Jiang}, Jiahao {Huang}, and Akiko {Aizawa}.
\newblock {JMedBench: A Benchmark for Evaluating Japanese Biomedical Large Language Models}.
\newblock \emph{arXiv e-prints}, art. arXiv:2409.13317, September 2024.
\newblock \doi{10.48550/arXiv.2409.13317}.

\bibitem[{Jin} et~al.(2020){Jin}, {Pan}, {Oufattole}, {Weng}, {Fang}, and {Szolovits}]{jin2020disease}
Di~{Jin}, Eileen {Pan}, Nassim {Oufattole}, Wei-Hung {Weng}, Hanyi {Fang}, and Peter {Szolovits}.
\newblock {What Disease does this Patient Have? A Large-scale Open Domain Question Answering Dataset from Medical Exams}.
\newblock \emph{arXiv e-prints}, art. arXiv:2009.13081, September 2020.
\newblock \doi{10.48550/arXiv.2009.13081}.

\bibitem[Jin et~al.(2019)Jin, Dhingra, Liu, Cohen, and Lu]{jin-etal-2019-pubmedqa}
Qiao Jin, Bhuwan Dhingra, Zhengping Liu, William Cohen, and Xinghua Lu.
\newblock {P}ub{M}ed{QA}: A dataset for biomedical research question answering.
\newblock In Kentaro Inui, Jing Jiang, Vincent Ng, and Xiaojun Wan, editors, \emph{Proceedings of the 2019 Conference on Empirical Methods in Natural Language Processing and the 9th International Joint Conference on Natural Language Processing (EMNLP-IJCNLP)}, pages 2567--2577, Hong Kong, China, November 2019. Association for Computational Linguistics.
\newblock \doi{10.18653/v1/D19-1259}.
\newblock URL \url{https://aclanthology.org/D19-1259/}.

\bibitem[Johri et~al.(2025)Johri, Jeong, Tran, Schlessinger, Wongvibulsin, Barnes, Zhou, Cai, Van~Allen, Kim, et~al.]{johri2025evaluation}
Shreya Johri, Jaehwan Jeong, Benjamin~A Tran, Daniel~I Schlessinger, Shannon Wongvibulsin, Leandra~A Barnes, Hong-Yu Zhou, Zhuo~Ran Cai, Eliezer~M Van~Allen, David Kim, et~al.
\newblock An evaluation framework for clinical use of large language models in patient interaction tasks.
\newblock \emph{Nature Medicine}, pages 1--10, 2025.

\bibitem[{Kanithi} et~al.(2024){Kanithi}, {Christophe}, {Pimentel}, {Raha}, {Saadi}, {Javed}, {Maslenkova}, {Hayat}, {Rajan}, and {Khan}]{kanithi2024medic}
Praveen~K {Kanithi}, Cl{\'e}ment {Christophe}, Marco~AF {Pimentel}, Tathagata {Raha}, Nada {Saadi}, Hamza {Javed}, Svetlana {Maslenkova}, Nasir {Hayat}, Ronnie {Rajan}, and Shadab {Khan}.
\newblock {MEDIC: Towards a Comprehensive Framework for Evaluating LLMs in Clinical Applications}.
\newblock \emph{arXiv e-prints}, art. arXiv:2409.07314, September 2024.
\newblock \doi{10.48550/arXiv.2409.07314}.

\bibitem[Kasai et~al.(2023)Kasai, Kasai, Sakaguchi, Yamada, and Radev]{kasai2023igakuqa}
Jungo Kasai, Yuhei Kasai, Keisuke Sakaguchi, Yutaro Yamada, and Dragomir Radev.
\newblock Evaluating gpt-4 and chatgpt on japanese medical licensing examinations, 2023.
\newblock URL \url{https://arxiv.org/abs/2303.18027}.

\bibitem[{Kim} et~al.(2025){Kim}, {Jeong}, {Chen}, {Li}, {Lu}, {Alhamoud}, {Mun}, {Grau}, {Jung}, {Gameiro}, {Fan}, {Park}, {Lin}, {Yoon}, {Yoon}, {Sap}, {Tsvetkov}, {Liang}, {Xu}, {Liu}, {McDuff}, {Lee}, {Park}, {Tulebaev}, and {Breazeal}]{kim2025medical}
Yubin {Kim}, Hyewon {Jeong}, Shan {Chen}, Shuyue~Stella {Li}, Mingyu {Lu}, Kumail {Alhamoud}, Jimin {Mun}, Cristina {Grau}, Minseok {Jung}, Rodrigo {Gameiro}, Lizhou {Fan}, Eugene {Park}, Tristan {Lin}, Joonsik {Yoon}, Wonjin {Yoon}, Maarten {Sap}, Yulia {Tsvetkov}, Paul {Liang}, Xuhai {Xu}, Xin {Liu}, Daniel {McDuff}, Hyeonhoon {Lee}, Hae~Won {Park}, Samir {Tulebaev}, and Cynthia {Breazeal}.
\newblock {Medical Hallucinations in Foundation Models and Their Impact on Healthcare}.
\newblock \emph{arXiv e-prints}, art. arXiv:2503.05777, February 2025.
\newblock \doi{10.48550/arXiv.2503.05777}.

\bibitem[{Luong} et~al.(2024){Luong}, {Zhang}, {Jie}, {Sun}, {Jin}, and {Li}]{luong2024reft}
Trung~Quoc {Luong}, Xinbo {Zhang}, Zhanming {Jie}, Peng {Sun}, Xiaoran {Jin}, and Hang {Li}.
\newblock {ReFT: Reasoning with Reinforced Fine-Tuning}.
\newblock \emph{arXiv e-prints}, art. arXiv:2401.08967, January 2024.
\newblock \doi{10.48550/arXiv.2401.08967}.

\bibitem[Mou et~al.(2024)Mou, Chen, Lode, Truhn, Sowe, and Decker]{mou2024radlink}
Yongli Mou, Hanbin Chen, Gwendolyn~Isabella Lode, Daniel Truhn, Sulayman Sowe, and Stefan Decker.
\newblock Radlink: Linking clinical entities from radiology reports.
\newblock In \emph{2024 2nd International Conference on Foundation and Large Language Models (FLLM)}, pages 443--449. IEEE, 2024.

\bibitem[Nori et~al.(2023)Nori, Lee, Zhang, Carignan, Edgar, Fusi, King, Larson, Li, Liu, Luo, McKinney, Ness, Poon, Qin, Usuyama, White, and Horvitz]{nori2023medprompt}
Harsha Nori, Yin~Tat Lee, Sheng Zhang, Dean Carignan, Richard Edgar, Nicolo Fusi, Nicholas King, Jonathan Larson, Yuanzhi Li, Weishung Liu, Renqian Luo, Scott~Mayer McKinney, Robert~Osazuwa Ness, Hoifung Poon, Tao Qin, Naoto Usuyama, Chris White, and Eric Horvitz.
\newblock Can generalist foundation models outcompete special-purpose tuning? case study in medicine.
\newblock \emph{arXiv preprint arXiv:2311.16452}, 2023.
\newblock URL \url{https://arxiv.org/abs/2311.16452}.

\bibitem[{Nori} et~al.(2023){Nori}, {Lee}, {Zhang}, {Carignan}, {Edgar}, {Fusi}, {King}, {Larson}, {Li}, {Liu}, {Luo}, {McKinney}, {Osazuwa Ness}, {Poon}, {Qin}, {Usuyama}, {White}, and {Horvitz}]{nori2023can}
Harsha {Nori}, Yin~Tat {Lee}, Sheng {Zhang}, Dean {Carignan}, Richard {Edgar}, Nicolo {Fusi}, Nicholas {King}, Jonathan {Larson}, Yuanzhi {Li}, Weishung {Liu}, Renqian {Luo}, Scott~Mayer {McKinney}, Robert {Osazuwa Ness}, Hoifung {Poon}, Tao {Qin}, Naoto {Usuyama}, Chris {White}, and Eric {Horvitz}.
\newblock {Can Generalist Foundation Models Outcompete Special-Purpose Tuning? Case Study in Medicine}.
\newblock \emph{arXiv e-prints}, art. arXiv:2311.16452, November 2023.
\newblock \doi{10.48550/arXiv.2311.16452}.

\bibitem[Nori et~al.(2024)Nori, Usuyama, King, McKinney, Fernandes, Zhang, and Horvitz]{nori2024medprompto1}
Harsha Nori, Naoto Usuyama, Nicholas King, Scott~Mayer McKinney, Xavier Fernandes, Sheng Zhang, and Eric Horvitz.
\newblock From medprompt to o1: Exploration of run-time strategies for medical challenge problems and beyond.
\newblock \emph{arXiv preprint arXiv:2411.03590}, 2024.
\newblock URL \url{https://arxiv.org/abs/2411.03590}.

\bibitem[OpenAI(2024)]{mmmlu}
OpenAI.
\newblock {openai/MMMLU}, 2024.
\newblock \url{https://huggingface.co/datasets/openai/MMMLU}.

\bibitem[{OpenAI} et~al.(2024){OpenAI}, {:}, {Hurst}, {Lerer}, {Goucher}, {Perelman}, {Ramesh}, {Clark}, {Ostrow}, {Welihinda}, {Hayes}, {Radford}, {Madry}, {Baker-Whitcomb}, {Beutel}, {Borzunov}, {Carney}, {Chow}, {Kirillov}, {Nichol}, {Paino}, {Renzin}, {Tachard Passos}, {Kirillov}, {Christakis}, {Conneau}, {Kamali}, {Jabri}, {Moyer}, {Tam}, {Crookes}, {Tootoochian}, {Tootoonchian}, {Kumar}, {Vallone}, {Karpathy}, {Braunstein}, {Cann}, {Codispoti}, {Galu}, {Kondrich}, {Tulloch}, {Mishchenko}, {Baek}, {Jiang}, {Pelisse}, {Woodford}, {Gosalia}, {Dhar}, {Pantuliano}, {Nayak}, {Oliver}, {Zoph}, {Ghorbani}, {Leimberger}, {Rossen}, {Sokolowsky}, {Wang}, {Zweig}, {Hoover}, {Samic}, {McGrew}, {Spero}, {Giertler}, {Cheng}, {Lightcap}, {Walkin}, {Quinn}, {Guarraci}, {Hsu}, {Kellogg}, {Eastman}, {Lugaresi}, {Wainwright}, {Bassin}, {Hudson}, {Chu}, {Nelson}, {Li}, {Shern}, {Conger}, {Barette}, {Voss}, {Ding}, {Lu}, {Zhang}, {Beaumont}, {Hallacy}, {Koch}, {Gibson}, {Kim}, {Choi}, {McLeavey}, {Hesse}, {Fischer},
  {Winter}, {Czarnecki}, {Jarvis}, {Wei}, {Koumouzelis}, {Sherburn}, {Kappler}, {Levin}, {Levy}, {Carr}, {Farhi}, {Mely}, {Robinson}, {Sasaki}, {Jin}, {Valladares}, {Tsipras}, {Li}, {Nguyen}, {Findlay}, {Oiwoh}, {Wong}, {Asdar}, {Proehl}, {Yang}, {Antonow}, {Kramer}, {Peterson}, {Sigler}, {Wallace}, {Brevdo}, {Mays}, {Khorasani}, {Petroski Such}, {Raso}, {Zhang}, {von Lohmann}, {Sulit}, {Goh}, {Oden}, {Salmon}, {Starace}, {Brockman}, {Salman}, {Bao}, {Hu}, {Wong}, {Wang}, {Schmidt}, {Whitney}, {Jun}, {Kirchner}, {Ponde de Oliveira Pinto}, {Ren}, {Chang}, {Chung}, {Kivlichan}, {O'Connell}, {O'Connell}, {Osband}, {Silber}, {Sohl}, {Okuyucu}, {Lan}, {Kostrikov}, {Sutskever}, {Kanitscheider}, {Gulrajani}, {Coxon}, {Menick}, {Pachocki}, {Aung}, {Betker}, {Crooks}, {Lennon}, {Kiros}, {Leike}, {Park}, {Kwon}, {Phang}, {Teplitz}, {Wei}, {Wolfe}, {Chen}, {Harris}, {Varavva}, {Lee}, {Shieh}, {Lin}, {Yu}, {Weng}, {Tang}, {Yu}, {Jang}, {Quinonero Candela}, {Beutler}, {Landers}, {Parish}, {Heidecke}, {Schulman},
  {Lachman}, {McKay}, {Uesato}, {Ward}, and {Kim}]{hurst2024gpt}
{OpenAI}, {:}, Aaron {Hurst}, Adam {Lerer}, Adam~P. {Goucher}, Adam {Perelman}, Aditya {Ramesh}, Aidan {Clark}, AJ~{Ostrow}, Akila {Welihinda}, Alan {Hayes}, Alec {Radford}, Aleksander {Madry}, Alex {Baker-Whitcomb}, Alex {Beutel}, Alex {Borzunov}, Alex {Carney}, Alex {Chow}, Alex {Kirillov}, Alex {Nichol}, Alex {Paino}, Alex {Renzin}, Alex {Tachard Passos}, Alexander {Kirillov}, Alexi {Christakis}, Alexis {Conneau}, Ali {Kamali}, Allan {Jabri}, Allison {Moyer}, Allison {Tam}, Amadou {Crookes}, Amin {Tootoochian}, Amin {Tootoonchian}, Ananya {Kumar}, Andrea {Vallone}, Andrej {Karpathy}, Andrew {Braunstein}, Andrew {Cann}, Andrew {Codispoti}, Andrew {Galu}, Andrew {Kondrich}, Andrew {Tulloch}, Andrey {Mishchenko}, Angela {Baek}, Angela {Jiang}, Antoine {Pelisse}, Antonia {Woodford}, Anuj {Gosalia}, Arka {Dhar}, Ashley {Pantuliano}, Avi {Nayak}, Avital {Oliver}, Barret {Zoph}, Behrooz {Ghorbani}, Ben {Leimberger}, Ben {Rossen}, Ben {Sokolowsky}, Ben {Wang}, Benjamin {Zweig}, Beth {Hoover}, Blake {Samic}, Bob
  {McGrew}, Bobby {Spero}, Bogo {Giertler}, Bowen {Cheng}, Brad {Lightcap}, Brandon {Walkin}, Brendan {Quinn}, Brian {Guarraci}, Brian {Hsu}, Bright {Kellogg}, Brydon {Eastman}, Camillo {Lugaresi}, Carroll {Wainwright}, Cary {Bassin}, Cary {Hudson}, Casey {Chu}, Chad {Nelson}, Chak {Li}, Chan~Jun {Shern}, Channing {Conger}, Charlotte {Barette}, Chelsea {Voss}, Chen {Ding}, Cheng {Lu}, Chong {Zhang}, Chris {Beaumont}, Chris {Hallacy}, Chris {Koch}, Christian {Gibson}, Christina {Kim}, Christine {Choi}, Christine {McLeavey}, Christopher {Hesse}, Claudia {Fischer}, Clemens {Winter}, Coley {Czarnecki}, Colin {Jarvis}, Colin {Wei}, Constantin {Koumouzelis}, Dane {Sherburn}, Daniel {Kappler}, Daniel {Levin}, Daniel {Levy}, David {Carr}, David {Farhi}, David {Mely}, David {Robinson}, David {Sasaki}, Denny {Jin}, Dev {Valladares}, Dimitris {Tsipras}, Doug {Li}, Duc~Phong {Nguyen}, Duncan {Findlay}, Edede {Oiwoh}, Edmund {Wong}, Ehsan {Asdar}, Elizabeth {Proehl}, Elizabeth {Yang}, Eric {Antonow}, Eric {Kramer}, Eric
  {Peterson}, Eric {Sigler}, Eric {Wallace}, Eugene {Brevdo}, Evan {Mays}, Farzad {Khorasani}, Felipe {Petroski Such}, Filippo {Raso}, Francis {Zhang}, Fred {von Lohmann}, Freddie {Sulit}, Gabriel {Goh}, Gene {Oden}, Geoff {Salmon}, Giulio {Starace}, Greg {Brockman}, Hadi {Salman}, Haiming {Bao}, Haitang {Hu}, Hannah {Wong}, Haoyu {Wang}, Heather {Schmidt}, Heather {Whitney}, Heewoo {Jun}, Hendrik {Kirchner}, Henrique {Ponde de Oliveira Pinto}, Hongyu {Ren}, Huiwen {Chang}, Hyung~Won {Chung}, Ian {Kivlichan}, Ian {O'Connell}, Ian {O'Connell}, Ian {Osband}, Ian {Silber}, Ian {Sohl}, Ibrahim {Okuyucu}, Ikai {Lan}, Ilya {Kostrikov}, Ilya {Sutskever}, Ingmar {Kanitscheider}, Ishaan {Gulrajani}, Jacob {Coxon}, Jacob {Menick}, Jakub {Pachocki}, James {Aung}, James {Betker}, James {Crooks}, James {Lennon}, Jamie {Kiros}, Jan {Leike}, Jane {Park}, Jason {Kwon}, Jason {Phang}, Jason {Teplitz}, Jason {Wei}, Jason {Wolfe}, Jay {Chen}, Jeff {Harris}, Jenia {Varavva}, Jessica~Gan {Lee}, Jessica {Shieh}, Ji~{Lin}, Jiahui
  {Yu}, Jiayi {Weng}, Jie {Tang}, Jieqi {Yu}, Joanne {Jang}, Joaquin {Quinonero Candela}, Joe {Beutler}, Joe {Landers}, Joel {Parish}, Johannes {Heidecke}, John {Schulman}, Jonathan {Lachman}, Jonathan {McKay}, Jonathan {Uesato}, Jonathan {Ward}, and Jong~Wook {Kim}.
\newblock {GPT-4o System Card}.
\newblock \emph{arXiv e-prints}, art. arXiv:2410.21276, October 2024.
\newblock \doi{10.48550/arXiv.2410.21276}.

\bibitem[Ouyang et~al.(2022)Ouyang, Wu, Jiang, Almeida, Wainwright, Mishkin, Zhang, Agarwal, Slama, Ray, et~al.]{ouyang2022training}
Long Ouyang, Jeffrey Wu, Xu~Jiang, Diogo Almeida, Carroll Wainwright, Pamela Mishkin, Chong Zhang, Sandhini Agarwal, Katarina Slama, Alex Ray, et~al.
\newblock Training language models to follow instructions with human feedback.
\newblock \emph{Advances in neural information processing systems}, 35:\penalty0 27730--27744, 2022.

\bibitem[{Pal} et~al.(2022){Pal}, {Umapathi}, and {Sankarasubbu}]{pal2022medmcqa}
Ankit {Pal}, Logesh~Kumar {Umapathi}, and Malaikannan {Sankarasubbu}.
\newblock {MedMCQA : A Large-scale Multi-Subject Multi-Choice Dataset for Medical domain Question Answering}.
\newblock \emph{arXiv e-prints}, art. arXiv:2203.14371, March 2022.
\newblock \doi{10.48550/arXiv.2203.14371}.

\bibitem[{Pang} et~al.(2024){Pang}, {Yuan}, {Cho}, {He}, {Sukhbaatar}, and {Weston}]{pang2024iterative}
Richard~Yuanzhe {Pang}, Weizhe {Yuan}, Kyunghyun {Cho}, He~{He}, Sainbayar {Sukhbaatar}, and Jason {Weston}.
\newblock {Iterative Reasoning Preference Optimization}.
\newblock \emph{arXiv e-prints}, art. arXiv:2404.19733, April 2024.
\newblock \doi{10.48550/arXiv.2404.19733}.

\bibitem[{Preferred Elements} et~al.(2024){Preferred Elements}, {:}, {Abe}, {Chubachi}, {Fujita}, {Hirokawa}, {Imajo}, {Kataoka}, {Komatsu}, {Mikami}, {Mogami}, {Murai}, {Nakago}, {Nishino}, {Ogawa}, {Okanohara}, {Ozaki}, {Sano}, {Suzuki}, {Xu}, and {Yanase}]{abe2024plamo}
{Preferred Elements}, {:}, Kenshin {Abe}, Kaizaburo {Chubachi}, Yasuhiro {Fujita}, Yuta {Hirokawa}, Kentaro {Imajo}, Toshiki {Kataoka}, Hiroyoshi {Komatsu}, Hiroaki {Mikami}, Tsuguo {Mogami}, Shogo {Murai}, Kosuke {Nakago}, Daisuke {Nishino}, Toru {Ogawa}, Daisuke {Okanohara}, Yoshihiko {Ozaki}, Shotaro {Sano}, Shuji {Suzuki}, Tianqi {Xu}, and Toshihiko {Yanase}.
\newblock {PLaMo-100B: A Ground-Up Language Model Designed for Japanese Proficiency}.
\newblock \emph{arXiv e-prints}, art. arXiv:2410.07563, October 2024.
\newblock \doi{10.48550/arXiv.2410.07563}.

\bibitem[{Qwen} et~al.(2024){Qwen}, {:}, {Yang}, {Yang}, {Zhang}, {Hui}, {Zheng}, {Yu}, {Li}, {Liu}, {Huang}, {Wei}, {Lin}, {Yang}, {Tu}, {Zhang}, {Yang}, {Yang}, {Zhou}, {Lin}, {Dang}, {Lu}, {Bao}, {Yang}, {Yu}, {Li}, {Xue}, {Zhang}, {Zhu}, {Men}, {Lin}, {Li}, {Tang}, {Xia}, {Ren}, {Ren}, {Fan}, {Su}, {Zhang}, {Wan}, {Liu}, {Cui}, {Zhang}, and {Qiu}]{yang2024qwen2}
{Qwen}, {:}, An~{Yang}, Baosong {Yang}, Beichen {Zhang}, Binyuan {Hui}, Bo~{Zheng}, Bowen {Yu}, Chengyuan {Li}, Dayiheng {Liu}, Fei {Huang}, Haoran {Wei}, Huan {Lin}, Jian {Yang}, Jianhong {Tu}, Jianwei {Zhang}, Jianxin {Yang}, Jiaxi {Yang}, Jingren {Zhou}, Junyang {Lin}, Kai {Dang}, Keming {Lu}, Keqin {Bao}, Kexin {Yang}, Le~{Yu}, Mei {Li}, Mingfeng {Xue}, Pei {Zhang}, Qin {Zhu}, Rui {Men}, Runji {Lin}, Tianhao {Li}, Tianyi {Tang}, Tingyu {Xia}, Xingzhang {Ren}, Xuancheng {Ren}, Yang {Fan}, Yang {Su}, Yichang {Zhang}, Yu~{Wan}, Yuqiong {Liu}, Zeyu {Cui}, Zhenru {Zhang}, and Zihan {Qiu}.
\newblock {Qwen2.5 Technical Report}.
\newblock \emph{arXiv e-prints}, art. arXiv:2412.15115, December 2024.
\newblock \doi{10.48550/arXiv.2412.15115}.

\bibitem[{Rafailov} et~al.(2023){Rafailov}, {Sharma}, {Mitchell}, {Ermon}, {Manning}, and {Finn}]{Rafailov2023}
Rafael {Rafailov}, Archit {Sharma}, Eric {Mitchell}, Stefano {Ermon}, Christopher~D. {Manning}, and Chelsea {Finn}.
\newblock {Direct Preference Optimization: Your Language Model is Secretly a Reward Model}.
\newblock \emph{arXiv e-prints}, art. arXiv:2305.18290, May 2023.
\newblock \doi{10.48550/arXiv.2305.18290}.

\bibitem[Saab et~al.(2024)Saab, Tu, Weng, Tanno, Stutz, Wulczyn, Zhang, Strother, Park, Vedadi, Chaves, Hu, Schaekermann, Kamath, Cheng, Barrett, Cheung, Mustafa, Palepu, McDuff, Hou, Golany, Liu, baptiste Alayrac, Houlsby, Tomasev, Freyberg, Lau, Kemp, Lai, Azizi, Kanada, Man, Kulkarni, Sun, Shakeri, He, Caine, Webson, Latysheva, Johnson, Mansfield, Lu, Rivlin, Anderson, Green, Wong, Krause, Shlens, Dominowska, Eslami, Chou, Cui, Vinyals, Kavukcuoglu, Manyika, Dean, Hassabis, Matias, Webster, Barral, Corrado, Semturs, Mahdavi, Gottweis, Karthikesalingam, and Natarajan]{saab2024medgemini}
Khaled Saab, Tao Tu, Wei-Hung Weng, Ryutaro Tanno, David Stutz, Ellery Wulczyn, Fan Zhang, Tim Strother, Chunjong Park, Elahe Vedadi, Juanma~Zambrano Chaves, Szu-Yeu Hu, Mike Schaekermann, Aishwarya Kamath, Yong Cheng, David G.~T. Barrett, Cathy Cheung, Basil Mustafa, Anil Palepu, Daniel McDuff, Le~Hou, Tomer Golany, Luyang Liu, Jean baptiste Alayrac, Neil Houlsby, Nenad Tomasev, Jan Freyberg, Charles Lau, Jonas Kemp, Jeremy Lai, Shekoofeh Azizi, Kimberly Kanada, SiWai Man, Kavita Kulkarni, Ruoxi Sun, Siamak Shakeri, Luheng He, Ben Caine, Albert Webson, Natasha Latysheva, Melvin Johnson, Philip Mansfield, Jian Lu, Ehud Rivlin, Jesper Anderson, Bradley Green, Renee Wong, Jonathan Krause, Jonathon Shlens, Ewa Dominowska, S.~M.~Ali Eslami, Katherine Chou, Claire Cui, Oriol Vinyals, Koray Kavukcuoglu, James Manyika, Jeff Dean, Demis Hassabis, Yossi Matias, Dale Webster, Joelle Barral, Greg Corrado, Christopher Semturs, S.~Sara Mahdavi, Juraj Gottweis, Alan Karthikesalingam, and Vivek Natarajan.
\newblock Capabilities of gemini models in medicine.
\newblock \emph{arXiv preprint arXiv:2404.18416}, 2024.
\newblock URL \url{https://arxiv.org/abs/2404.18416}.

\bibitem[Singhal et~al.(2025)Singhal, Tu, Gottweis, Sayres, Wulczyn, Amin, Hou, Clark, Pfohl, Cole-Lewis, et~al.]{singhal2025medpalm}
Karan Singhal, Tao Tu, Juraj Gottweis, Rory Sayres, Ellery Wulczyn, Mohamed Amin, Le~Hou, Kevin Clark, Stephen~R Pfohl, Heather Cole-Lewis, et~al.
\newblock Toward expert-level medical question answering with large language models.
\newblock \emph{Nature Medicine}, pages 1--8, 2025.

\bibitem[Sukeda(2024)]{sukeda2024}
Issey Sukeda.
\newblock Development and bilingual evaluation of japanese medical large language model within reasonably low computational resources.
\newblock \emph{arXiv preprint arXiv:2409.11783}, 2024.
\newblock URL \url{https://arxiv.org/abs/2409.11783}.

\bibitem[{Sukeda} et~al.(2024){Sukeda}, {Kishikawa}, and {Kodera}]{sukeda202470b}
Issey {Sukeda}, Risa {Kishikawa}, and Satoshi {Kodera}.
\newblock {70B-parameter large language models in Japanese medical question-answering}.
\newblock \emph{arXiv e-prints}, art. arXiv:2406.14882, June 2024.
\newblock \doi{10.48550/arXiv.2406.14882}.

\bibitem[Tu et~al.(2025)Tu, Schaekermann, Palepu, Saab, Freyberg, Tanno, Wang, Li, Amin, Cheng, et~al.]{tu2025towards}
Tao Tu, Mike Schaekermann, Anil Palepu, Khaled Saab, Jan Freyberg, Ryutaro Tanno, Amy Wang, Brenna Li, Mohamed Amin, Yong Cheng, et~al.
\newblock Towards conversational diagnostic artificial intelligence.
\newblock \emph{Nature}, pages 1--9, 2025.

\bibitem[Van~Veen et~al.(2024)Van~Veen, Van~Uden, Blankemeier, Delbrouck, Aali, Bluethgen, Pareek, Polacin, Reis, Seehofnerov{\'a}, et~al.]{van2024adapted}
Dave Van~Veen, Cara Van~Uden, Louis Blankemeier, Jean-Benoit Delbrouck, Asad Aali, Christian Bluethgen, Anuj Pareek, Malgorzata Polacin, Eduardo~Pontes Reis, Anna Seehofnerov{\'a}, et~al.
\newblock Adapted large language models can outperform medical experts in clinical text summarization.
\newblock \emph{Nature medicine}, 30\penalty0 (4):\penalty0 1134--1142, 2024.

\bibitem[Xie et~al.(2024)Xie, Wu, Tu, Yang, Zhao, Zong, Jin, Xie, and Zhou]{xie2024}
Yunfei Xie, Juncheng Wu, Haoqin Tu, Siwei Yang, Bingchen Zhao, Yongshuo Zong, Qiao Jin, Cihang Xie, and Yuyin Zhou.
\newblock A preliminary study of o1 in medicine: Are we closer to an ai doctor?
\newblock \emph{arXiv preprint arXiv:2409.15277}, 2024.
\newblock URL \url{https://arxiv.org/abs/2409.15277}.

\bibitem[Zhang et~al.(2023)Zhang, Chen, Jiang, Yu, Chen, Li, Chen, Wu, Zhang, Xiao, Wan, Wang, and Li]{zhang2023huatuogpt}
Hongbo Zhang, Junying Chen, Feng Jiang, Fei Yu, Zhihong Chen, Jianquan Li, Guiming Chen, Xiangbo Wu, Zhiyi Zhang, Qingying Xiao, Xiang Wan, Benyou Wang, and Haizhou Li.
\newblock Huatuogpt, towards taming language model to be a doctor.
\newblock \emph{arXiv preprint arXiv:2305.15075}, 2023.
\newblock URL \url{https://arxiv.org/abs/2305.15075}.

\end{thebibliography}
\bibliographystyle{plainnat}

\newpage
\appendix
\onecolumn
\section{Appendix}
\counterwithin{table}{section}
\setcounter{table}{0}

\begin{table}[htbp]
\caption{Example of human-generated explanatory data}
\label{appendix_table}
\begin{center}
\begin{small}
\begin{tabular}{p{1.5cm} p{13cm}}
\toprule
\textbf{Key} & \textbf{Value} \\
\midrule
\textit{problem} &
A 45-year-old woman. She was noted to have liver dysfunction 
in two consecutive yearly company health examinations and came 
to our hospital. She does not drink alcohol. Her height is 158 cm 
and her weight is 46 kg. Abdominal examination reveals no 
abnormalities, and the liver is not palpable. Blood chemistry 
findings: fasting blood glucose 86 mg/dL, total protein 7.6 g/dL, 
AST 62 U/L, ALT 106 U/L, ALP 200 U/L (reference range $\leq$260), 
${\rm\gamma}$-GTP 35 U/L (reference range 8–50). Which autoantibodies are 
useful for diagnosis? Choose two. 
\\
\hline
\textit{choices} & 
\begin{tabular}[t]{@{}l@{}}
a. Anti-nuclear antibody (ANA) \\
b. Anti-ENA antibody \\
c. Anti-Jo-1 antibody \\
d. Anti-smooth muscle antibody (ASMA) \\
e. Anti-mitochondrial antibody (AMA)
\end{tabular} \\
\hline
\textit{explanation} & 
A young female with no history of alcohol consumption presenting with liver function abnormalities. While AST and ALT show mild elevations, ALP and ${\rm\gamma}$-GTP remain within normal limits, suggesting autoimmune hepatitis. 
a and d are correct: anti-nuclear antibodies and anti-smooth 
muscle antibodies are positive in autoimmune hepatitis.
b Anti-ENA antibodies are elevated in systemic lupus erythematosus (SLE), 
Sjögren’s syndrome, mixed connective tissue disease (MCTD), etc. Note that 
patients with autoimmune hepatitis may also have comorbidities such as 
Sjögren’s syndrome or Hashimoto’s disease. *Currently, anti-ENA antibody 
testing is not covered by insurance and is not generally performed. 
c Anti-Jo-1 antibody is elevated in polymyositis/dermatomyositis. 
e Anti-mitochondrial antibody is elevated in primary biliary cholangitis (PBC). 
\\
\hline
\textit{answer} & a, d \\
\bottomrule
\end{tabular}
\end{small}
\end{center}
\end{table}

\end{document}